\documentclass{article}

\PassOptionsToPackage{numbers}{natbib}

\usepackage[final]{neurips_distshift_2021}

\usepackage[utf8]{inputenc} 
\usepackage[T1]{fontenc}    
\usepackage{hyperref}       
\usepackage{url}            
\usepackage{booktabs}       
\usepackage{amsfonts}       
\usepackage{nicefrac}       
\usepackage{microtype}      
\usepackage{xcolor}         
\usepackage{multirow}

\usepackage{amsmath,amsfonts,bm}

{}
{}
{}
{}

\def\eqref#1{equation~\ref{#1}}

\def\1{\bm{1}}

\DeclareMathAlphabet{\mathsfit}{\encodingdefault}{\sfdefault}{m}{sl}
\SetMathAlphabet{\mathsfit}{bold}{\encodingdefault}{\sfdefault}{bx}{n}

\usepackage{graphicx}

\title{Improving Baselines in the Wild}

\author{%
  Kazuki Irie$^{1}$ ~ Imanol Schlag$^{1}$ ~ R\'obert Csord\'as$^{1}$ ~ J\"urgen Schmidhuber$^{1,2}$\\
  $^1$The Swiss AI Lab, IDSIA, University of Lugano (USI) \& SUPSI, Lugano, Switzerland \\
  $^2$King Abdullah University of Science and Technology (KAUST), Thuwal, Saudi Arabia \\
  \texttt{\{kazuki, imanol, robert, juergen\}@idsia.ch} \\
}

\begin{document}

\maketitle

\begin{abstract}
We share our experience with the recently released 
WILDS benchmark, a collection of ten datasets dedicated to developing models and training strategies
which are robust to domain shifts.
Several experiments yield a couple of critical observations which we believe are of general interest for any future work on WILDS.
Our study focuses on two datasets: iWildCam and FMoW.
We show that
(1) Conducting separate cross-validation for each evaluation metric is crucial for both datasets,
(2) A weak correlation between validation and test performance
might make model development difficult for iWildCam,
(3) Minor changes in the training of hyper-parameters improve the baseline by a relatively large margin (mainly on FMoW),
(4) There is a strong correlation between certain domains and certain target labels (mainly on iWildCam).
To the best of our knowledge, 
no prior work on these datasets has reported 
these observations despite their obvious importance.
Our code is public.\footnote{\url{https://github.com/kazuki-irie/fork--wilds-public}}
\end{abstract}

\section{Introduction}
\label{sec:intro}
It goes without saying that common benchmark datasets and solid baselines are
essential ingredients for developing new models and correctly measuring progress in machine learning.
Key datasets such as
ImageNet \citep{dengDSLL009} in image recognition,
Switchboard \citep{godfreyHM92} in automatic speech recognition,
or Penn Treebank \citep{MarcusSM94, mikolov2011extensions} in language modelling,
to name only a few,
have played a crucial role in demonstrating the
impressive performance of neural networks (NNs) especially in the 2010s.
They have contributed to gradually marking the progress obtained by new techniques and models.
Equally importantly, strong baselines are crucial to properly measure progress.
In some cases, by revisiting some old or standard baseline configurations, we end up with surprisingly good results, including results of LSTMs for language modelling \citep{MelisDB18}, ResNets for vision \citep{bello2021revisiting},
or Transformers for systematic generalization \citep{csordas2021devil}.

Nowadays, few researchers are surprised by NNs performing very well
on some in-domain data distribution,
and there has been increasing interest in
developing models that are robust to domain shifts \citep{abnar2021gradual, shi2021gradient, malinin2021shifts, koh2021wilds}.
Here we focus on the recently proposed benchmark for evaluating domain robust systems, WILDS \citep{koh2021wilds}, and we share our empirical experience with two datasets of WILDS, iWildCam and FMoW, as well as their baseline models.
A handful of experiments exhibit important, previously unreported facets of the data and the baselines.
Our main findings can be summarised as follows:
(1) Conducting separate cross-validation for each evaluation metric is crucial for both datasets,
(2) A weak correlation between validation and test performance
might make model development difficult (iWildCam),
(3) Minor changes in the training hyper-parameters improve the baseline by a relatively large margin (mainly on FMoW),
(4) There is a strong correlation between certain domains and certain target labels (mainly on iWildCam).
Any practitioner should be aware of these aspects when developing new ideas based on these
two datasets, as well as on other datasets of WILDS.

\section{Datasets and Experimental Settings}
\label{sec:data}
Here we briefly summarize essential properties of the datasets
used in our experiments.
For further information, we refer to the original paper by \citet{koh2021wilds}.

\paragraph{iWildCam2020-WILDS.}
\textsc{iWildCam2020-WILDS} (iWildCam for short) is a variant of the iWildCam 2020 Competition Dataset \citep{beery2020iwildcam}.
The task is classification of photos of 182 different animal species from various camera traps.
The camera traps define various \textit{domains} of this dataset.
The training set contains about 130~K photos taken by 243 camera traps.
The out-of-distribution validation and test sets consist of photos from 32 and 48 different camera traps
which are not part of the training domains.
Following \citet{koh2021wilds}, we report the
classification accuracy, as well as the macro F1 score which
\citet{koh2021wilds} set as the most relevant evaluation metric.

\paragraph{FMoW-WILDS.}
FMoW-WILDS (FMoW for short) is a variant of the functional map of the world dataset
\citep{christieFWM18}.
This is also an image classification task
using satellite images containing one of 62 building or land classes.
There are two attributes for grouping these images: years and geographical regions (Africa, the Americas, Oceania, Asia, or Europe).
The training data consist of images taken before 2013,
the validation from 2013 to 2015, and the test set from 2016.
\citet{koh2021wilds} set the worst region accuracy (WR)
as the main evaluation metric, and also
report the total accuracy (TA).

\paragraph{Experimental settings.}
Unless otherwise stated, we use all default settings
provided by the publicly available official codebase for WILDS \citep{koh2021wilds}.

\begin{table}[t]
\caption{Performance on \textbf{iWildCam}.
``CV'' denotes the cross-validation criterion (on the OOD validation set) used to select the best checkpoint.
* denotes information which is not available in the original paper.
Following \citet{koh2021wilds}, we report mean and std
over 3 training runs.}
\label{tab:main_iwild}
\setlength{\tabcolsep}{0.3em}
\begin{center}
\noindent\makebox[\textwidth]{
\begin{tabular}{lccrrrrrrr}
\toprule
Model & CV  &   \multicolumn{4}{c}{Accuracy (AC)}  & \multicolumn{4}{c}{Macro F1 (F1)}   \\ \cmidrule(l){3-6} \cmidrule(l){7-10}
& &    \multicolumn{2}{c}{Valid}  &  \multicolumn{2}{c}{Test} & \multicolumn{2}{c}{Valid} & \multicolumn{2}{c}{Test} \\
& &  \multicolumn{1}{c}{ID} & \multicolumn{1}{c}{OOD} & \multicolumn{1}{c}{ID} & \multicolumn{1}{c}{OOD} & \multicolumn{1}{c}{ID} & \multicolumn{1}{c}{OOD} & \multicolumn{1}{c}{ID} & \multicolumn{1}{c}{OOD} \\ \midrule
 \citet{koh2021wilds} & \multicolumn{1}{c}{*} & 82.5 (0.8) & 62.7 (2.4) & 75.7 (0.3) & 71.6 (2.5) & \textbf{48.8} (2.5) & 37.4 (1.7) & 47.0 (1.4) & 31.0 (1.3) \\ \midrule
Our run & F1 & 82.7 (1.1) & 61.7 (0.4) & 74.9 (0.2) & 69.9 (0.1) & 47.9 (0.6) & 36.2 (0.4) & 45.3 (0.4) & 30.2 (1.2) \\
 & AC & 82.7 (0.1) & 64.7 (0.4) & 75.6 (0.3) & \textbf{71.9} (1.8) & 45.0 (4.2) & 33.6 (3.3) & 41.1 (5.3) & 27.2 (3.4) \\ \midrule
+ freq ckpt  & F1 & 82.5 (0.8) & 64.1 (1.7) & \textbf{76.2} (0.1) & 69.0 (0.3) & 46.7 (1.0) & \textbf{38.3} (0.9) & \textbf{47.9} (2.1) & \textbf{32.1} (1.2) \\
 & AC & \textbf{82.6} (0.7) & \textbf{66.6} (0.4) & 75.8 (0.4) & 68.6 (0.3) & 46.2 (0.9) & 36.6 (2.1) & 44.9 (0.4) & 28.7 (2.0) \\
 \bottomrule
\end{tabular}
} 
\end{center}
\end{table}

\section{Core findings}
\label{sec:core}
In this section, we highlight our findings
on \textsc{iWildCam2020-wilds} (iWildCam for short) and \textsc{FMoW-wilds} (FMoW for short).
We note that all our results are produced using
the official baseline code provided by \citet{koh2021wilds} with small modifications specified below.

\paragraph{Metric-wise cross-validation is crucial.}
In both iWildCam and FMoW, the performance of models is
measured on two or more evaluation metrics (one of them is considered to be the ``main'' metric).
E.g., for iWildCam,
the classification accuracy (AC) and macro F1 (F1) are reported on in-domain
(ID) and out-of-domain (OOD) subsets of validation and test sets.
Assuming that we mainly care about the OOD performance,
this results in two metrics to monitor during the model development: OOD validation AC and F1.
Technically, this implies a necessity for
conducting cross-validation and checkpoint tracking
separately for each metric.
We note that such a metric-wise cross-validation is not part
of the official setting \citep{koh2021wilds}.
Furthermore, we observe that in the official setting,
the cross-validation is carried out only at the end of each epoch.
We increase the cross-validation frequency (to every 1000 steps)
such that we do not miss good checkpoints between epochs.
Table \ref{tab:main_iwild}
shows the performance of different checkpoints selected using two different CV criteria within the same training runs on iWildCam.
The first thing to notice here is that simply introducing
checkpoint tracking between epochs can improve the baseline performance by a non-negligible margin.
In Table \ref{tab:main_iwild}, if we compare ``Our run''
which uses the default setting by \citet{koh2021wilds}
and ``+ freq ckpt'' which does cross validation
every 1000 training steps:
the OOD validation accuracy improves from 64.7\% to 66.6\%
and the F1 from 36.2\% to 38.3\%.

The impact of metric-wise CV is also relatively large:
In our best configuration (``+ freq ckpt'' in Table \ref{tab:main_iwild}), the AC-CV checkpoint (i.e.~the best checkpoint found by cross validation based on OOD validation accuracy) achieves an OOD validation accuracy of 66.6\% vs. 64.1\% for the F1-CV one.
Similarly, the F1-CV model achieves an OOD validation F1 score of 38.3\% compared to 36.6\% for the AC-CV one.

We stress that these performance gaps are obtained without any
technical changes on the algorithmic level,
revealing that very careful tuning is needed
to compare models on this dataset.
Strictly speaking, we should also do separate cross-validations on the ID validation metrics to report and compare the ID performance.
We omit this here as our results on the OOD metrics already clearly exhibit the issue.
While our goal is not to develop separate models for different metrics,
we observe how crucial it is to report the cross-validation setting for fair comparisons between 
different approaches.

\paragraph{Correlation between validation and test performance is weak.}
Another important observation we drawn from Table \ref{tab:main_iwild} is that
despite a relatively large performance gap between
the F1-CV and AC-CV models in terms of OOD validation accuracy
(64.1 vs.~66.6\%), their OOD test accuracy is similar (69.0 vs.~68.6\%).
This trend is less visible for FMoW (Table \ref{tab:main_fmow}, batch size 64), but still: our WR-CV model outperforms the official
baseline on the WR validation accuracy (50.2 vs.~48.9\%) while it underperforms
on the corresponding WR test accuracy (31.4 vs.~32.3\%).
This is actually not surprising as the validation and test sets
are also OOD to each other, hence improvements on the validation set
do not necessarily transfer to the test set.
However, this is problematic since we only have access to the OOD validation performance while developing models.
The common practice of selecting the checkpoint based on a validation
set (also recommended by the original work \citep{koh2021wilds};
in the paragraph on avoiding overfitting to the test sets)
might not work here: we thus have no reliable reference metric for developing new models.
A similar problem regarding the lack of useful validation sets has been recently pointed out \citep{csordas2021devil} in
the context of systematic generalization \citep{fodor1988connectionism}.
This problem seems even harder for general domain shifts:
unlike in the case of systematic generalization,
there is no a priori way of controlling the degree of "domain shifts" systematically.
This calls for discussing the construction of \textit{useful} validation sets when studying OOD generalization.

\begin{table}[t]
\caption{Performance on \textbf{FMoW}. ``Batch'' column indicates the batch size.
``CV'' denotes the Cross-validation Criterion (on the OOD validation set).
$^*$not mentioned in the corresponding paper.
Following \citet{koh2021wilds}, we report mean and std dev
over 3 training runs.}
\label{tab:main_fmow}
\begin{center}
\noindent\makebox[\textwidth]{
\begin{tabular}{lccrrrrrrrr}
\toprule
Batch & CV &   \multicolumn{4}{c}{Total Accuracy (TA)}  & \multicolumn{4}{c}{Worst Region Accuracy (WR)}   \\ \cmidrule(l){3-6} \cmidrule(l){7-10}
 & &     \multicolumn{2}{c}{Valid}  &  \multicolumn{2}{c}{Test} & \multicolumn{2}{c}{Valid} & \multicolumn{2}{c}{Test} \\
& &  \multicolumn{1}{c}{IID} & \multicolumn{1}{c}{OOD} & \multicolumn{1}{c}{IID} & \multicolumn{1}{c}{OOD} & \multicolumn{1}{c}{IID} & \multicolumn{1}{c}{OOD} & \multicolumn{1}{c}{IID} & \multicolumn{1}{c}{OOD} \\ \midrule
 {64 \citep{koh2021wilds}} & \multicolumn{1}{c}{*} & 61.2 (0.5) & 59.5 (0.4) & 59.7 (0.7) & 53.0 (0.6) & 59.2 (0.7) & 48.9 (0.6) & 58.3 (0.9) & 32.3 (1.3) \\  \midrule
64 & TA  & 60.8 (0.2) & 59.3 (0.3) & 59.7 (0.1) & 52.6 (0.4) & 58.7 (0.3) & 47.6 (1.9) & 58.2 (0.3) & 30.9 (2.6) \\ 
 & WR & 59.8 (0.8) & 57.9 (0.5) & 58.7 (0.5) & 51.9 (0.4) & 57.8 (1.0) & 50.2 (1.2) & 57.0 (0.5) & 31.4 (0.9) \\ 
 \midrule
32 & TA  & 61.2 (0.1) & 59.8 (0.1) & 60.1 (0.1) & 53.3 (0.2) & 59.2 (0.3) & 48.1 (1.1) & 58.5 (0.5) & 33.4 (0.1) \\ 
 & WR & 60.4 (0.9) & 58.8 (0.6) & 59.3 (0.6) & 52.5 (0.5) & 58.3 (0.9) & 51.0 (0.9) & 57.7 (0.5) & 33.6 (1.6) \\ \midrule
20  & TA  & 62.1 (0.1) & 60.3 (0.2) & 60.7 (0.1) & 53.8 (0.3) & 60.4 (0.2) & 49.6 (0.1) & 59.0 (0.1) & 33.8 (0.5) \\ 
 & WR & 61.3 (0.3) & 58.9 (0.6) & 59.9 (0.4) & 52.5 (0.2) & 59.0 (0.3) & 52.4 (0.8) & 58.5 (0.5) & 34.4 (0.3) \\ \midrule
+ higher lr & TA & \textbf{64.0} (0.1) & \textbf{62.1} (0.0) & \textbf{62.3} (0.4) & \textbf{55.6} (0.2) & \textbf{62.3} (0.4) & 51.4 (1.3) & \textbf{61.1} (0.6) & 34.2 (1.2) \\ 
 & WR & 63.9 (0.2) & \textbf{62.1} (0.0) & \textbf{62.3} (0.2) & \textbf{55.6} (0.2) & 62.2 (0.5) & \textbf{52.5} (1.0) & 60.9 (0.6) & \textbf{34.8} (1.5) \\
\midrule \midrule
Fish \citep{shi2021gradient} & \multicolumn{1}{c}{*} & \multicolumn{1}{c}{*} & 57.8 (0.2) & \multicolumn{1}{c}{*} & 51.8 (0.3) & \multicolumn{1}{c}{*}  & 49.5 (2.3) & \multicolumn{1}{c}{*} & 34.6 (0.2) \\
\bottomrule
\end{tabular}
} 
\end{center}
\end{table}

\paragraph{There is sub-optimality in the baseline.}
Another crucial observation regarding FMoW is the sub-optimality
of the baseline setting.
While keeping all configurations equal to the original configuration \citep{koh2021wilds},
we modify the training batch size in the FMoW baseline.
As is shown in Table \ref{tab:main_fmow}, simply by reducing the batch size from
64 to 20, the best/worst region accuracy (WR; the main evaluation metric for FMoW) improves
from 50.2\% to 52.4\% on the OOD validation set,
and from 31.4\% to 34.4\% on the OOD test set.
By further modifying the learning rate from 1e-4 to 3e-4
and increasing the checkpoint frequency to be every 200 steps instead of 1000, we finally obtain a worst region accuracy of 34.8\% (rows ``+ higher lr'').
We note that these final OOD test accuracies we obtain are competitive
compared to the number reported\footnote{These numbers were taken from \url{https://wilds.stanford.edu/leaderboard/} as recommended by \citet{shi2021gradient} (personal communication).} by \citet{shi2021gradient}, 34.6\%, achieved using
a technique specifically designed for the domain shift problem (a method called Fish; last row in Table \ref{tab:main_fmow}).
Such ``trivial'' improvements over the baseline call for further
tuning of the baseline settings before developing and evaluating
new models on this dataset.

\begin{figure}[h]
    \begin{center}
        \includegraphics[width=0.5\columnwidth]{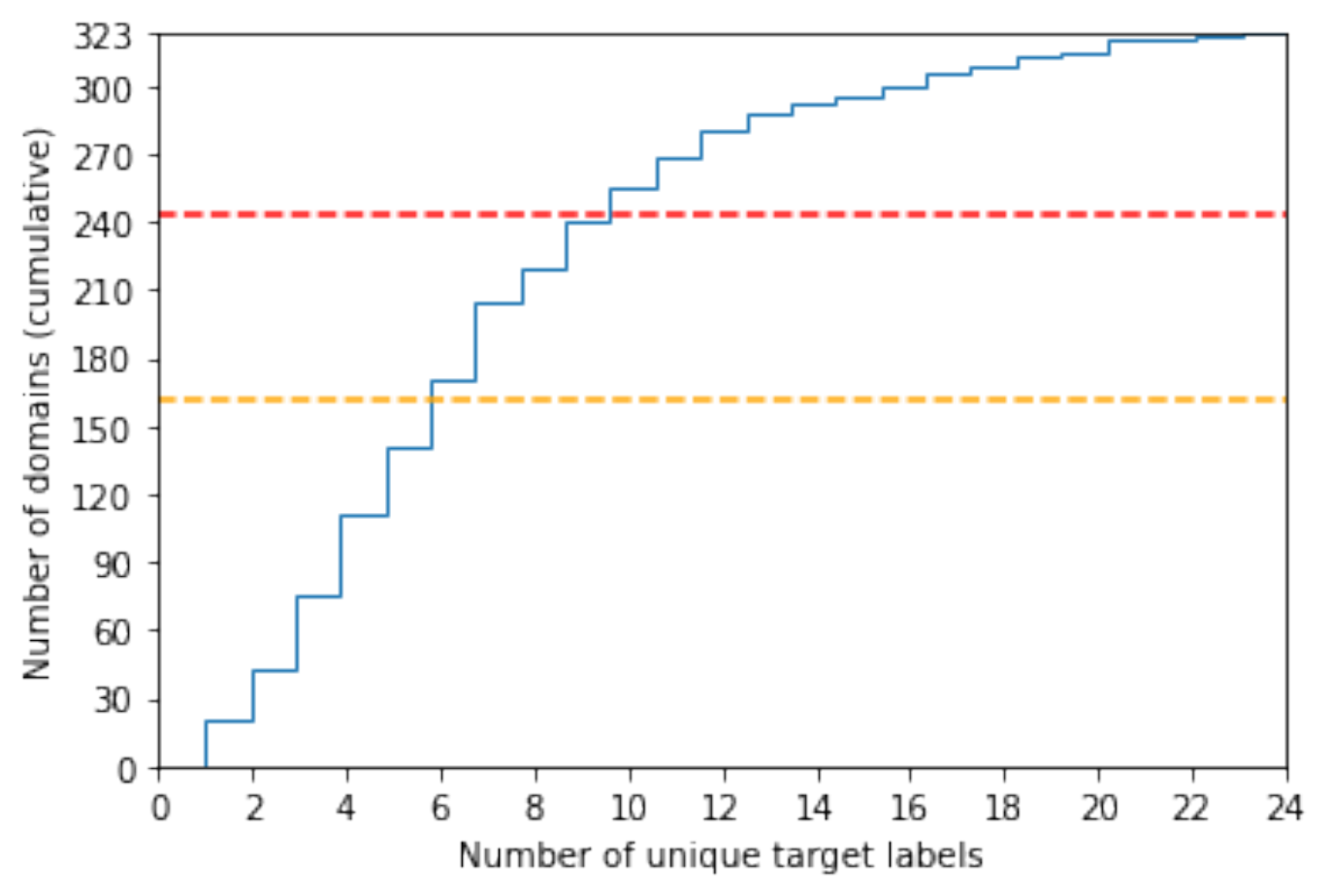}
        \caption{
Cumulative histogram over the number of domains
with the corresponding number of unique target class labels within the domain.
The yellow line indicates the coverage of 50\% of the total number of domains, while the red one indicates 75\%.}
        \label{fig:iwild}
    \end{center}
\end{figure}

\paragraph{Correlation between domains and target labels is rather strong.}
Our last observation to be shared here is that
certain domains have a bias towards certain target labels.
Each image in iWildCam is labeled with a target class label (one of 182 animal species) and a domain label (one of 323 camera traps).
For each domain, we count the number of unique class labels
covered within the domain
i.e., if this number is 1 for a certain domain, all images belonging to that domain have the same target class label.
Figure \ref{fig:iwild} shows the cumulative histogram of the corresponding statistics.
As can be seen, the number of unique class labels (x-axis) vary
only from 1 to 24 (while the maximum is 182)
and the 50\% of the domains contain less than 5 distinct classes (indicated by the yellow line).
This number improves to 10 distinct classes only when we take 75\% of the domains (red line).
Overall, the diversity of target class labels is very limited within each domain.
This bias can potentially be problematic when developing training strategies which make use of some grouping of datapoints by the domain label.
A number of techniques has been proposed to exploit domain information \citep{sunS16, arjovsky2019invariant, huNSS18}
for domain robust learning.
Interestingly, none of these techniques have
resulted in consistent improvements over the basic empirical risk minimisation baseline \citep{koh2021wilds}.
We point out these previously unreported important statistical properties although further analysis is required to draw conclusions regarding the causal relationship between these two observations. 

\section{Conclusion}
\label{sec:ccl}
To properly evaluate new machine learning methods and measure progress,
it is crucial to start from strong and well-established baselines.
We reported a couple of important observations from our experiments with the baseline settings of WILDS.
The latter can be improved by simply tweaking a few configurations (frequent metric-wise cross-validation) and one hyper-parameter for FMoW (batch size).
Our observations seem to indicate that a systematic study of baseline
configurations is necessary before starting the development of new models based on WILDS.
We hope that practitioners will find our observations useful,
and take them into account for future work on WILDS.

\begin{ack}
This research was partially funded by ERC Advanced grant no: 742870, project AlgoRNN,
and by Swiss National Science Foundation grant no: 200021\_192356, project NEUSYM.
This work was partially supported by
computational resources at the CSCS Swiss
National Supercomputing Centre, project d115.
We thank NVIDIA Corporation for donating several DGX
machines, and IBM for donating a Minsky machine.
\end{ack}

\bibliography{main}

\begin{thebibliography}{17}
\providecommand{\natexlab}[1]{#1}
\providecommand{\url}[1]{\texttt{#1}}
\expandafter\ifx\csname urlstyle\endcsname\relax
  \providecommand{\doi}[1]{doi: #1}\else
  \providecommand{\doi}{doi: \begingroup \urlstyle{rm}\Url}\fi

\bibitem[Deng et~al.(2009)Deng, Dong, Socher, Li, Li, and
  Fei{-}Fei]{dengDSLL009}
Jia Deng, Wei Dong, Richard Socher, Li{-}Jia Li, Kai Li, and Li~Fei{-}Fei.
\newblock Image{N}et: {A} large-scale hierarchical image database.
\newblock In \emph{Proc. {IEEE} Conf. on Computer Vision and Pattern
  Recognition (CVPR)}, pages 248--255, Miami, {FL}, {USA}, 2009.

\bibitem[Godfrey et~al.(1992)Godfrey, Holliman, and McDaniel]{godfreyHM92}
John~J. Godfrey, Edward Holliman, and Jane McDaniel.
\newblock {SWITCHBOARD:} telephone speech corpus for research and development.
\newblock In \emph{Proc. {IEEE} Int. Conf. on Acoustics, Speech and Signal
  Processing (ICASSP)}, pages 517--520, San Francisco, {CA}, USA, March 1992.

\bibitem[Marcus et~al.(1993)Marcus, Santorini, and Marcinkiewicz]{MarcusSM94}
Mitchell~P. Marcus, Beatrice Santorini, and Mary~Ann Marcinkiewicz.
\newblock Building a large annotated corpus of english: The penn treebank.
\newblock \emph{Comput. Linguistics}, 19\penalty0 (2):\penalty0 313--330, 1993.

\bibitem[Mikolov et~al.(2011)Mikolov, Kombrink, Burget, Cernocky, and
  Khudanpur]{mikolov2011extensions}
Tomas Mikolov, Stefan Kombrink, Lukas Burget, Jan~H Cernocky, and Sanjeev
  Khudanpur.
\newblock Extensions of recurrent neural network language model.
\newblock In \emph{Proc. {IEEE} Int. Conf. on Acoustics, Speech and Signal
  Processing (ICASSP)}, pages 5528--5531, Prague, Czech Republic, May 2011.

\bibitem[Melis et~al.(2018)Melis, Dyer, and Blunsom]{MelisDB18}
G{\'{a}}bor Melis, Chris Dyer, and Phil Blunsom.
\newblock On the state of the art of evaluation in neural language models.
\newblock In \emph{6th International Conference on Learning Representations,
  {ICLR} 2018, Vancouver, BC, Canada, April 30 - May 3, 2018, Conference Track
  Proceedings}, Vancouver, Canada, 2018.

\bibitem[Bello et~al.(2021)Bello, Fedus, Du, Cubuk, Srinivas, Lin, Shlens, and
  Zoph]{bello2021revisiting}
Irwan Bello, William Fedus, Xianzhi Du, Ekin~D Cubuk, Aravind Srinivas,
  Tsung-Yi Lin, Jonathon Shlens, and Barret Zoph.
\newblock Revisiting {R}es{N}ets: Improved training and scaling strategies.
\newblock In \emph{Proc. Advances in Neural Information Processing Systems
  (NeurIPS)}, Virtual only, 2021.

\bibitem[Csord\'as et~al.(2021)Csord\'as, Irie, and
  Schmidhuber]{csordas2021devil}
R\'obert Csord\'as, Kazuki Irie, and J\"urgen Schmidhuber.
\newblock The devil is in the detail: Simple tricks improve systematic
  generalization of transformers.
\newblock In \emph{Proc. Conf. on Empirical Methods in Natural Language
  Processing (EMNLP)}, Punta Cana, Dominican Republic, November 2021.

\bibitem[Abnar et~al.(2021)Abnar, Berg, Ghiasi, Dehghani, Kalchbrenner, and
  Sedghi]{abnar2021gradual}
Samira Abnar, Rianne van~den Berg, Golnaz Ghiasi, Mostafa Dehghani, Nal
  Kalchbrenner, and Hanie Sedghi.
\newblock Gradual domain adaptation in the wild: When intermediate
  distributions are absent.
\newblock \emph{Preprint arXiv:2106.06080}, 2021.

\bibitem[Shi et~al.(2021)Shi, Seely, Torr, Siddharth, Hannun, Usunier, and
  Synnaeve]{shi2021gradient}
Yuge Shi, Jeffrey Seely, Philip~HS Torr, N~Siddharth, Awni Hannun, Nicolas
  Usunier, and Gabriel Synnaeve.
\newblock Gradient matching for domain generalization.
\newblock \emph{Preprint arXiv:2104.09937}, 2021.

\bibitem[Malinin et~al.(2021)Malinin, Band, Chesnokov, Gal, Gales, Noskov,
  Ploskonosov, Prokhorenkova, Provilkov, Raina, et~al.]{malinin2021shifts}
Andrey Malinin, Neil Band, German Chesnokov, Yarin Gal, Mark~JF Gales, Alexey
  Noskov, Andrey Ploskonosov, Liudmila Prokhorenkova, Ivan Provilkov, Vatsal
  Raina, et~al.
\newblock Shifts: A dataset of real distributional shift across multiple
  large-scale tasks.
\newblock \emph{Preprint arXiv:2107.07455}, 2021.

\bibitem[Koh et~al.(2021)Koh, Sagawa, Xie, Zhang, Balsubramani, Hu, Yasunaga,
  Phillips, Gao, Lee, et~al.]{koh2021wilds}
Pang~Wei Koh, Shiori Sagawa, Sang~Michael Xie, Marvin Zhang, Akshay
  Balsubramani, Weihua Hu, Michihiro Yasunaga, Richard~Lanas Phillips, Irena
  Gao, Tony Lee, et~al.
\newblock Wilds: A benchmark of in-the-wild distribution shifts.
\newblock In \emph{Proc. Int. Conf. on Machine Learning (ICML)}, pages
  5637--5664, Virtual only, 2021.

\bibitem[Beery et~al.(2020)Beery, Agarwal, Cole, and
  Birodkar]{beery2020iwildcam}
Sara Beery, Arushi Agarwal, Elijah Cole, and Vighnesh Birodkar.
\newblock The iwildcam 2020 competition dataset.
\newblock \emph{Fine-Grained Visual Categorization Workshop at CVPR}, 2020.

\bibitem[Christie et~al.(2018)Christie, Fendley, Wilson, and
  Mukherjee]{christieFWM18}
Gordon Christie, Neil Fendley, James Wilson, and Ryan Mukherjee.
\newblock Functional map of the world.
\newblock In \emph{Proc. {IEEE} Conf. on Computer Vision and Pattern
  Recognition (CVPR)}, pages 6172--6180, Salt Lake City, UT, USA, June 2018.

\bibitem[Fodor et~al.(1988)Fodor, Pylyshyn, et~al.]{fodor1988connectionism}
Jerry~A Fodor, Zenon~W Pylyshyn, et~al.
\newblock Connectionism and cognitive architecture: A critical analysis.
\newblock \emph{Cognition}, 28\penalty0 (1-2):\penalty0 3--71, 1988.

\bibitem[Sun and Saenko(2016)]{sunS16}
Baochen Sun and Kate Saenko.
\newblock Deep {CORAL:} correlation alignment for deep domain adaptation.
\newblock In \emph{Computer Vision - {ECCV} Workshops}, pages 443--450,
  Amsterdam, The Netherlands, October 2016.

\bibitem[Arjovsky et~al.(2019)Arjovsky, Bottou, Gulrajani, and
  Lopez-Paz]{arjovsky2019invariant}
Martin Arjovsky, L{\'e}on Bottou, Ishaan Gulrajani, and David Lopez-Paz.
\newblock Invariant risk minimization.
\newblock \emph{Preprint arXiv:1907.02893}, 2019.

\bibitem[Hu et~al.(2018)Hu, Niu, Sato, and Sugiyama]{huNSS18}
Weihua Hu, Gang Niu, Issei Sato, and Masashi Sugiyama.
\newblock Does distributionally robust supervised learning give robust
  classifiers?
\newblock In \emph{Proc. Int. Conf. on Machine Learning (ICML)}, pages
  2034--2042, Stockholm, Sweden, July 2018.

\end{thebibliography}
\bibliographystyle{unsrtnat}

\clearpage
\appendix

\end{document}